\documentclass[lettersize,journal]{IEEEtran}
\usepackage{amsmath,amsfonts}
\usepackage{array}
\usepackage[caption=false,font=normalsize,labelfont=sf,textfont=sf]{subfig}
\usepackage{textcomp}
\usepackage{stfloats}
\usepackage{url}
\usepackage{verbatim}
\usepackage{graphicx}
\usepackage{eso-pic}
\def\BibTeX{{\rm B\kern-.05em{\sc i\kern-.025em b}\kern-.08em
    T\kern-.1667em\lower.7ex\hbox{E}\kern-.125emX}}
\usepackage{balance}

\usepackage{algorithm}
\usepackage{booktabs}
\usepackage{multirow}
\usepackage{algorithm}
\usepackage{algpseudocode}

\usepackage{hyperref}
\hypersetup{
    hidelinks
}
\usepackage{xcolor}
\definecolor{orcidlogocol}{HTML}{A6CE39}
\usepackage{orcidlink}

\begin{document}

\AddToShipoutPictureFG*{%
  \AtPageUpperLeft{%
    \put(\LenToUnit{0.5\paperwidth},\LenToUnit{-0.28in}){%
      \makebox(0,0)[t]{%
        \parbox{\textwidth}{\centering
          This work has been submitted to the IEEE for possible publication. Copyright may be transferred without notice, after which this version
          may no longer be accessible.}%
      }%
    }%
  }%
}


\title{AI Control Scientist: LLM-driven Agentic System for Automated Control Design}
\author{Anonymized Authors}

\author{Haiteng~Wang~\orcidlink{0000-0002-7316-3607},~\IEEEmembership{Member,~IEEE,}
        Weihao~Li, 
        Jing~Zhang~\orcidlink{0000-0002-1480-227X},~\IEEEmembership{Student~Member,~IEEE,}
        and~Lei~Ren~\orcidlink{0000-0001-6346-6930},~\IEEEmembership{Senior~Member,~IEEE}}
        

\maketitle

\begin{abstract}
Control system design is critical for modern industry, such as chemical process temperature regulation and aero-engine control. However,traditional control design workflows rely heavily on expert knowledge and extensive manual parameter tuning, resulting in limited efficiency and scalability. To this end, this paper proposes AI Control Scientist (AICS), the first large language model (LLM)-driven agent capable of automatically generating optimized controller from language design requirements. Specifically, a Task Modeling Agent interprets user requirements to engineering constraints; a Controller Design Agent generate candidate controller structures and executable code; and a Parameter Tuning Agent refine controller parameters under closed-loop performance criteria.  Experiments demonstrate that the proposed agentic system can automatically generate multiple representative control systems, outperforms existing automated baselines in both design success rate and optimization efficiency. This work has the potential to transform control system design from human-driven to agent-driven, paving the way for model predictive control and other advanced control systems design.
\end{abstract} 


\section{Introduction}
\IEEEPARstart{C}{ontrol} system design is the cornerstone of modern engineering science, underpinning a wide range of applications such as process manufacturing \cite{lu2023observer}, aerospace systems \cite{stevens2015aircraft}, autonomous driving \cite{chen2023milestones, gao2026soad},  and robotic dynamics \cite{siciliano2009robotics}. With the rapid advancement of Industry 4.0 \cite{lasi2014industry}, modern industrial systems is undergoing a paradigm shift from conventional automation toward fully autonomous operation \cite{kusiak2018smart}.In this landscape, industrial foundation models and industrial agents have emerged as pivotal accelerators, driving cross-domain intelligence and self-optimizing capabilities across manufacturing and process ecosystems \cite{ren2025industrial}. In this context, the ability to efficiently and accurately design robust and optimal control strategies has become a key determinant of competitiveness in high-end manufacturing and intelligent systems.

However, as system complexity increases, traditional control design approaches, such as heuristic tuning rules\cite{Ziegler1942OptimumSF, o2009handbook}, rely heavily on iterative manual trial-and-error procedures. These methods are often labor-intensive, time-consuming, and difficult to scale to high-dimensional systems with strong couplings, uncertainties, and multiple performance constraints. More importantly, valuable expert knowledge accumulated through practice is typically fragmented and difficult to translate into reusable automated workflows.

Recently, large language models (LLMs) have demonstrated remarkable capabilities in reasoning, code generation, and autonomous decision-making. Their success in automated programming, exemplified by systems such as AlphaCode \cite{li2022competition} and AgentCoder \cite{huang2023agentcoder}, has opened a promising avenue for AI-driven engineering design. Concurrently, the paradigm of foundation models has been comprehensively extended into industrial sectors, fostering specialized industrial foundation models capable of handling complex physical laws and multi-modal industrial modalities \cite{ren2025foundation}. Nevertheless, directly applying general-purpose code generation models such as Codex \cite{Chen2021EvaluatingLL} and CodeLlama \cite{codellama} to control engineering remains highly challenging. Unlike conventional software generation, where syntactic correctness and input-output consistency are typically sufficient, control-oriented code generation must simultaneously satisfy physical laws, closed-loop stability, numerical precision, and safety constraints\cite{brunke2022safe}. As a result, several fundamental barriers still hinder the deployment of LLMs in complex industrial control tasks.

\begin{figure}[htbp]
    \centering
    \includegraphics[width=\linewidth]{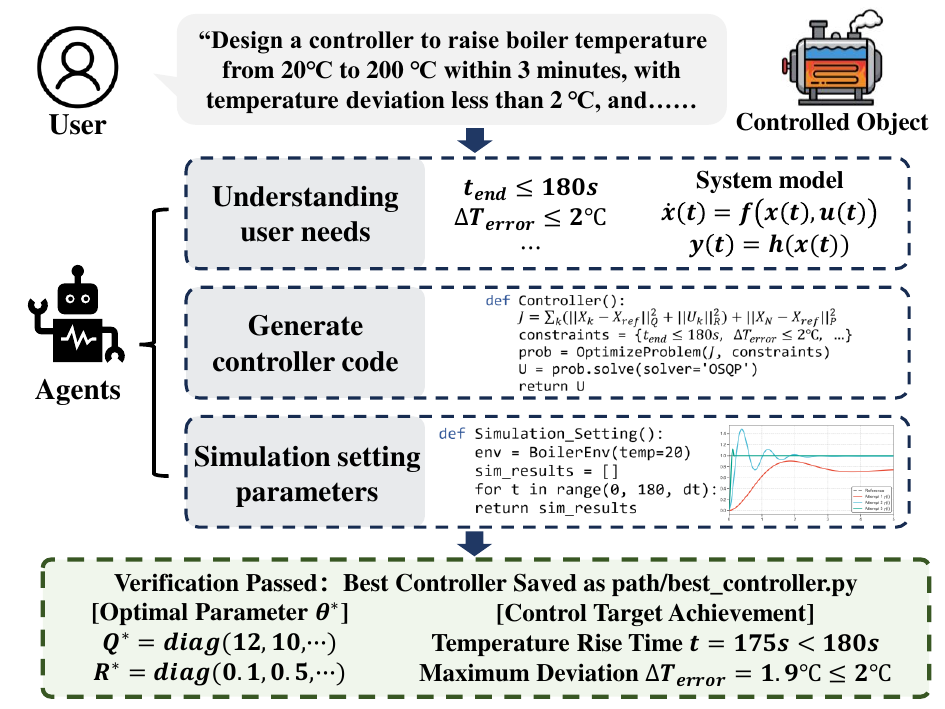}
    \caption{End-to-end execution example of the proposed LLM-driven agentic system for an industrial boiler control task. The system automatically transforms natural-language engineering requirements into rigorous mathematical constraints and outputs an optimized control strategy validated through closed-loop simulation.}
    \label{fig:intro_demo}
\end{figure}

\textbf{NLP-to-Control Structural Constraint Gap.}  
Industrial control problems are inherently structured, involving state variables, dynamic equations, constraints, and performance objectives \cite{kirk2004optimal}. In contrast, user requirements are routinely expressed in ambiguous natural language, such as "avoid overshoot" or "improve disturbance rejection". These descriptions implicitly encode control objectives and constraints but do not provide explicit mathematical formulations. Existing LLMs still struggle to reliably mapping such high-level, unstructured language specifications into rigorous optimization objectives or control constraints \cite{zhang2026optiverse, huang2025llms, song2026safety}.

\textbf{Difficulty of LLMs to Synthesize Logic-Compliant Control Code.}  
Many industrial controllers require reasoning over domain-specific physical structures. Examples include multi-loop PID coordination, model predictive control (MPC) with hard constraints \cite{garcia1989model}, and nonlinear feedback compensation. Effective controller synthesis depends on understanding stability margins, actuator saturation, coupling effects, and dynamic trade-offs. General-purpose LLMs, however, often lack grounded physical reasoning capabilities \cite{srivastava2023beyond} and therefore fail to generate control architectures that are both theoretically sound and practically deployable \cite{liu2023your}.

\textbf{Control Parameter Optimization Bottleneck.}
Even when a feasible controller structure is obtained, high-performance control critically depends on accurate parameter tuning, such as PID gains, observer poles, and MPC weighting matrices. Existing LLMs are known to suffer from numerical hallucination and unreliable arithmetic reasoning \cite{shao2025benford, imani2023mathprompter}, making them unsuitable for fine-grained quantitative optimization. Consequently, practitioners still rely on repeated simulation-based manual tuning by domain experts, which significantly limits automation efficiency.

To address the above challenges, this paper proposes a LLM-driven agentic system for automated control system design. Aligning with the paradigm of modern industrial agents \cite{ren2026industrial}, the key idea is to emulate the iterative workflow of experienced control engineers through a role-specialized multi-agent architecture. Specifically, a Task Modeling Agent (TMA) interprets user requirements and constructs system models together with engineering constraints; a Controller Design Agent (CDA) embeds control knowledge to generate candidate controller structures and executable code; and a Parameter Tuning Agent (PTA) invokes external numerical solvers to refine controller parameters under closed-loop performance criteria. Through iterative collaboration and feedback, the proposed system enables end-to-end automation from natural-language specifications to validated control implementations.

Experiments on multiple representative control systems demonstrate that the proposed system outperforms existing automated baselines in both design success rate and optimization efficiency, while achieving performance comparable to that of domain experts. These results suggest a scalable and generalizable paradigm toward AI-native control engineering.

The main contributions of this paper are summarized as follows:
\begin{itemize}
    \item \textbf{LLM-driven agentic system:} We establish a LLM-based multi-agent control design system. Rather than relying on a single-agent paradigm, the system employs three collaborative agents for automated control design: (1) a Task Modeling Agent for system identification and control-oriented modeling; (2) a Controller Design Agent for generating candidate controller structures and executable code; and (3) a Parameter Tuning Agent for parameter tuning under closed-loop performance criteria. The system enables a closed-loop workflow for end-to-end automated control design, effectively replacing traditional manual programming and significantly reducing the heavy reliance on expert knowledge.
    \item \textbf{Logic-Guided Controller Generation Paradigm:} Moving beyond random code generation, we introduce a logic-grounded synthesis paradigm. By explicitly  inferencing on critical system characteristics (e.g., time-delays, non-minimum phase zeros, and nonlinear coupling), the system generates a structured Design Blueprint ($B_{logic}$) to systematically guide the generation of controller code.
    \item \textbf{LLM-Driven Planner-Solver Optimization:} To avoid the limitations of numerical hallucination, we propose a planner-solver optimization mechanism for high-precision parameter tuning. The LLM acts as a high-level Planner to perform meta-reasoning for inferring optimization bounds and search priors, which are then passed to external numerical solvers to ensure numerical rigor, robust stability, and safety under practical constraints.
\end{itemize}

\section{Related Work}
\subsection{Large Language Models for Code Generation}
The advancement of Large Language Models (LLMs) has catalyzed a paradigm shift in software engineering, moving from basic code completion to autonomous agents capable of program synthesis and iterative refinement\cite{austin2021program}.
Crucial to this evolution is the emergence of closed-loop reflection mechanisms, which empower LLMs to evaluate their own outputs and systematically refine behaviors without explicit retraining \cite{shinn2023reflexion, madaan2023self}.
Expanding upon basic verbal reflection, recent methodologies focus on advanced planning trajectories. Techniques such as Tree-of-Thought (ToT)\cite{yao2023tree} and Monte Carlo Tree Search (MCTS)\cite{antoniades2025swe} view code generation as a non-linear state-space search. These reasoning-centric approaches significantly mitigate the risk of catastrophic planning failures in complex programmatic synthesis.

To bridge the semantic gap between textual patterns and functional correctness, CodeRL+\cite{jiang2025coderl+} integrates execution semantics alignment into reinforcement learning pipelines, allowing models to infer variable-level trajectories from execution feedback and significantly improving pass rates on competitive programming tasks. Frameworks like InterCode\cite{NEURIPS2023_4b175d84} have further standardized this interactive process by framing code generation as a reinforcement learning environment where code acts as actions and compiler feedback provides deterministic observations.
Beyond single-agent systems, multi-agent architectures such as MetaGPT\cite{hong2024metagpt} and ChatDev\cite{chatdev} simulate the collaborative workflows of software companies. By encoding Standardized Operating Procedures (SOPs) into prompt sequences, these systems assign specialized roles to different agents to ensure consistency and minimize cascading hallucinations in complex development projects.
Expanding this generative capability to physical domains, recent advances also leverage large-generative frameworks for industrial time series \cite{ren2025aigc}, incorporating frequency-aware models like MetaIndux-TS \cite{wang2025metaindux} to precisely model complex industrial dynamics.

\subsection{Large Language Models for Automated Control Design}
In the domain of control engineering, LLMs are being leveraged to bridge the gap between high-level natural language requirements and rigorous mathematical specifications. Code-as-Policies\cite{10160591} introduced the "Code Agents" paradigm, leveraging LLMs to directly transform natural-language instructions into executable control programs in Python or C++. 
This iterative code-generation loop has been further extended by incorporating environmental feedback and persistent skill libraries to achieve long-horizon task execution\cite{wang2023voyager}.

ControlAgent\cite{guo2024controlagent} emulates the iterative design process of practicing engineers by utilizing specialized agents for task distribution and a dedicated Python computation agent to perform precise frequency-domain evaluations and parameter refinements. Building on this, the AgenticControl\cite{Narimani2025AgenticControlAA} framework employs a structured six-agent architecture to systematically refine controller parameters across scenarios of increasing complexity, successfully handling nonlinear dynamics and parametric uncertainties.
Similarly, frameworks such as LLM-MPC \cite{maher2025llmpc} leverage LLMs to dynamically translate high-level goals into mathematical objective functions and operational constraints for Model Predictive Control, balancing flexible intent-parsing with numerical stability guarantees.
For safety-critical systems, Agents4PLC \cite{11410524}addresses the lack of formal guarantees in LLM-generated code by introducing a multi-agent system for closed-loop Programmable Logic Controller (PLC) code generation and formal verification. Furthermore, recent empirical studies on PID tuning\cite{11091752} indicate that models like GPT-4o can iteratively optimize controller gains through structured prompt engineering, achieving performance comparable to or exceeding traditional analytical methods.

\section{Method}
\subsection{Framework}
\begin{figure*}[htbp] 
    \centering
    \includegraphics[width=\textwidth]{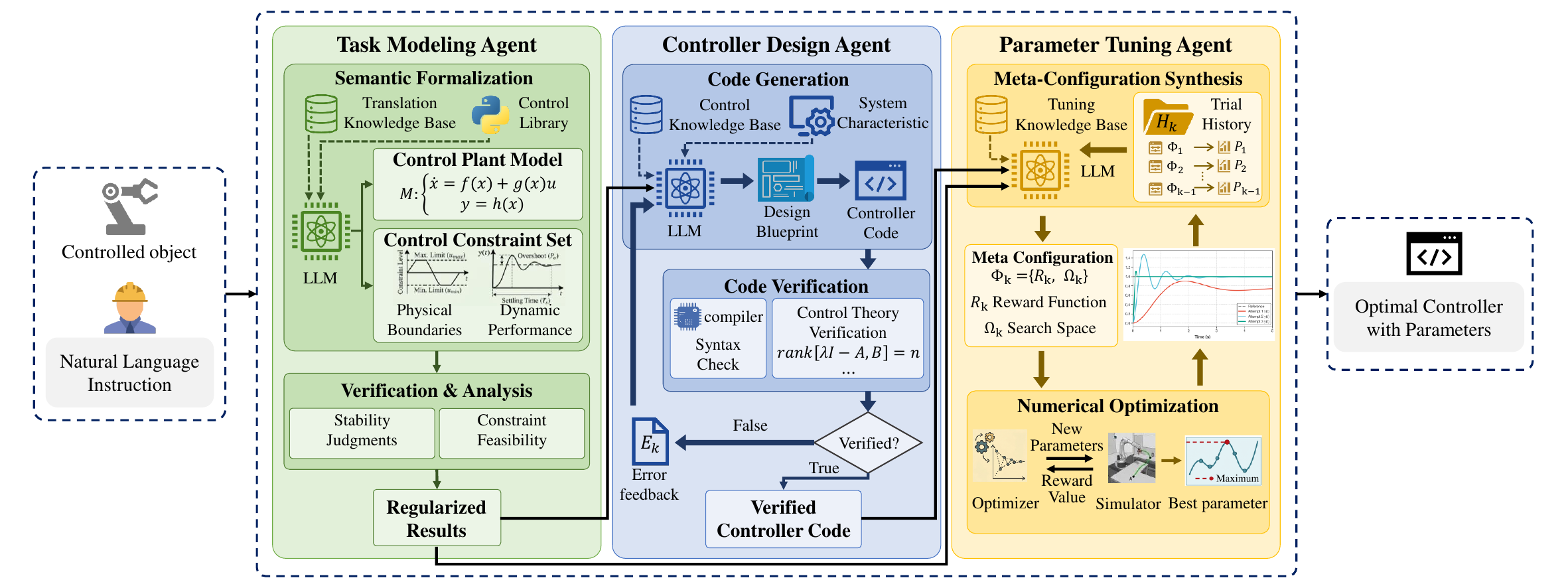} 
    \caption{The overall architecture of the proposed Multi-Agent System. It consists of a Task Modeling Agent for semantic formalization that translates ambiguous natural-language requirements into rigorous mathematical constraints, a Controller Design Agent for for logic-grounded code generation that synthesizes theoretically sound controller structures based on plant characteristics, and a Parameter Tuning Agent for precision optimization that leverages external numerical solvers to refine control gains under closed-loop stability criteria.}
    \label{fig:overall_framework}
\end{figure*}

In this study, we propose a large language model (LLM)-driven agentic system capable of automatically generating optimized control system. As illustrated in Fig.~\ref{fig:overall_framework}, our core aim is to address the challenge that single LLMs struggle to directly generate precise and stable controller for complex dynamic systems. By deeply integrating the logical reasoning capabilities of LLMs with the high-precision computation of numerical optimizers, this system leverages closed-loop feedback mechanisms among multiple agents to achieve fully automated design, from policy planning to the generation of expert-level, high-performance controllers.

Fundamentally, control system design is a high-dimensional, constrained, non-convex optimization problem. Defining the state space $X \subset \mathbb{R}^n$ and the input space $U \subset \mathbb{R}^m$, the global optimization objective of this framework is to find the optimal control law mapping $\pi^*: X \rightarrow U$ and its parameter set $\theta^*$ to minimize the comprehensive cost functional $J$:
$$ (\pi^*, \theta^*) = \mathop{\arg\min}_{\pi, \theta} \mathbb{E}_{x_0} \left[ J(x(\cdot), u(\cdot); \theta) \right] $$
$$ \text{s.t.} \quad \dot{x}(t) = f(x(t), u(t)), \quad c(x(t), u(t)) \le 0 $$
where $f(\cdot)$ is the system evolution function, and $c(\cdot)$ represents the physical and performance constraints.

To solve this complex problem, we design a sequential pipeline of "Semantic Formalization—Code Generation—Parameter Tuning," decoupling the aforementioned intractable global optimization problem into a multi-stage workflow executed by three agents. Given a natural language instruction set $L_{input}$:
$$ S_{spec} = TMA(L_{input}) $$
$$ C_{verified} = CDA(S_{spec}) $$
$$ (C^*, \theta^*) = PTA(C_{verified}, S_{spec}) $$
where $C$ is the code implementation of the control law $\pi$. The specific workflow of the system comprises the following three key stages:

\textbf{1. Semantic Formalization:} The workflow begins with the TMA receiving the user's natural language control objectives. This agent first eliminates semantic ambiguities by referencing a translation knowledge base, converting vague descriptions into deterministic system parameters and performance constraints. It then utilizes the conditional verification and system analysis tools of the Control standard function library to output a set of Regularized Results containing mathematical definitions and physical boundaries, providing an unambiguous input interface for downstream stages.

\textbf{2. Code Generation:} The CDA receives the formalized results, determines the optimal control strategy based on the control knowledge base, and generates the controller code. This stage employs a "Inference-Synthesis" pipeline, where the agent first identifies critical system features (e.g., time-delays, coupling) to establish a Design Rationale. A verify tool then checks both syntax and control-theoretic integrity, detected errors will be fed back for iterative refactoring until a verified controller code is finalized.

\textbf{3. Parameter Tuning:} Finally, the PTA takes over the verified code and constructs a dynamic closed loop within a Simulator. Unlike traditional direct parameter tuning by LLMs, this agent utilizes the reasoning capability of the LLM to define the reward function and parameter search range of the optimization problem according to the tuning knowledge base. This guides the optimization algorithm to search efficiently within the parameter space, ultimately outputting the optimal controller parameter configuration that satisfies all design metrics.

Through the above process, this framework achieves the end-to-end automated design of control systems driven by natural language, ensuring the theoretical correctness and engineering feasibility of the design results.

\subsection{Task Modeling Agent (TMA)}

As the front-end interface of the system, the TMA is responsible for the semantic understanding and strict mathematical formalization of the problem. The core task of this agent is to map unstructured text $L_{input}$ into a structured specification set $S_{spec}$. This set is defined as a tuple containing the controlled plant's dynamic model $M$ and the control constraint set $C_{constraint}$:
$$ S_{spec} = \{M, C_{constraint}\} = TMA(L_{input} | K_{perceive}) $$

The specific workflow of the TMA comprises the following two core subtasks:

\textbf{1. Semantic Formalization:} The agent leverages a translation knowledge base to resolve semantic ambiguities in the natural language input, precisely mapping unstructured engineering descriptions into a deterministic system dynamic model $M$ and a set of performance constraints $C_{constraint}$. For general systems, the model is formalized as:
$$ M : \begin{cases} \dot{x}(t) = f(x(t), u(t)) \\ y(t) = h(x(t)) \end{cases} $$
Simultaneously, it converts control requirements described in natural language into strict numerical boundaries. The constraint set is divided into physical execution boundaries $C_{phys}$ and dynamic performance metrics $C_{perf}$:
$$
\begin{aligned}
C_{phys} = \Big\{ & u(t) \in [u_{min}, u_{max}], \\
                            & \dot{u}(t) \in [\Delta u_{min}, \Delta u_{max}] \Big\}
\end{aligned}
$$
$$
\begin{aligned}
C_{perf} = \Big\{ & \sup_{t} y(t) \le y_{ref}(1 + \sigma_{max}), \\
                            & ||y(t) - y_{ref}|| \le e_{ss} \ \forall t \ge t_s \Big\}
\end{aligned}
$$

\textbf{2. System Verification and Analysis:} After completing the semantic formalization, the agent invokes the Control standard function library to perform verification and analysis on the extracted mathematical parameters. For Linear Time-Invariant (LTI) systems, the agent converts them into the standard constant-coefficient matrix form $\dot{x}(t) = Ax(t) + Bu(t)$, and conducts stability judgments based on the system order and open-loop pole locations . Concurrently, the verification module conducts a physical feasibility check on $C_{phys}$ and $C_{perf}$ to eliminate logical conflicts. Ultimately, it outputs a set of regularized results $S_{spec}$ containing rigorous mathematical definitions and physical boundaries, providing an unambiguous mathematical Feasible Region for subsequent steps.

\subsection{Controller Design Agent (CDA)}

The CDA focuses on the generation of controller code. Instead of the simplistic "text-to-code" generation approach, this module adopts a System-Characteristic-Driven Design Paradigm. The core philosophy is to ensure that the control logic is not merely a generic template but is explicitly tailored to the underlying physical properties of the plant. 
The agent’s workflow is structured into two fundamental stages:

\textbf{1. Characteristic-Driven Code Generation:} 
The agent first performs inference on the plant's dynamical properties to synthesize a design blueprint ($B_{logic}$), which serves as the theoretical blueprint for the controller.

The agent maps the formalized specification $S_{spec}$ to a set of key dynamical descriptors $D = \{ \tau, \zeta, \sigma, \dots \}$, representing time-delays, damping ratios, and singular values.
Then, Using the control knowledge base $K_{ctrl}$, the agent determines the optimal control law $\pi$. This is formulated as a two-step hierarchical inference:
$$ B_{logic} = \text{LLM}(S_{spec} \mid D, K_{ctrl}) $$
$$ c_{k} = \text{LLM}(B_{logic}, S_{spec} \mid K_{ctrl}) $$

By conditioning the code generation $c_k$ on $B_{logic}$, the agent ensures that the resulting implementation is not a stochastic guess but a deterministic consequence of the identified system characteristics.

\textbf{2. Backward Correction:} The generated code $c_k$ is fed into the verification tool $V(\cdot)$. This tool not only performs basic syntax checking $V_{syn}$, but also incorporates a control theory verification engine $V_{theory}$. By utilizing underlying theories such as controllability criteria, it preemptively intercepts invalid logic that violates control theories.

If $V(c_k) = V_{syn} \wedge V_{theory} = \text{False}$, the verification tool captures the exception stack and theory-violation information, encoding them into an error feedback operator $E_k$, which is then fed back to the LLM to generate a new controller:
$$c_{k+1} = \text{LLM}( S_{spec}, c_k, E_k | K_{ctrl} )$$
This Markov Decision Process iterates continuously until $V(c_k) = \text{True}$, outputting a verified controller $C_{verified}$.

\subsection{Parameter Tuning Agent (PTA)}

The PTA is responsible for solving the high-dimensional, non-convex parameter optimization problem. Addressing the pain point of hallucinations and numerical inaccuracies when pure LLMs directly generate specific numerical parameters, this agent adopts a Meta-Optimization paradigm characterized by "decoupling of symbolic reasoning and numerical computation."

Let $k$ be the meta-optimization decision round. The optimization configuration space $\Phi_k$ is defined as the set of the parameter search domain $\Omega_k$ and the search reward function $R_k(\cdot)$: $\Phi_k = \{ \Omega_k, R_k \}$.

\textbf{1. History-Driven Meta-Configuration Synthesis:} Instead of directly guessing control parameters, the LLM acts as a policy generator. It analyzes the nonlinear mapping relationship between the reward function structures and the final performance metrics $P^*$ within the trial history repository $H_k = \{ (\Phi_0, P^*_0), \dots, (\Phi_{k-1}, P^*_{k-1}) \}$. Through in-context learning, the LLM dynamically adjusts reward function $R_k$  or the search space $\Omega_k$:
$$\Phi_k = \text{LLM}(C_{verified}, S_{spec}, H_k | K_{tune})$$

\textbf{2. Numerical Optimization Engine:} Under the given meta-configuration $\Phi_k$, the underlying numerical optimizer takes over the search process. The optimizer constructs a Surrogate Model of the objective function and performs efficient sampling within $\Omega_k$ by maximizing the Acquisition Function ($\alpha(\theta)$):
$$ \theta_{t+1} = \mathop{\arg\max}_{\theta \in \Omega_k} \alpha(\theta; D_{1:t}) $$
Ultimately, within a limited simulation budget, it obtains the optimal parameter vector $\theta^*_k$ that maximizes the expected reward:
$$ \theta^*_k = \mathop{\arg\max}_{\theta \in \Omega_k} \mathbb{E}_{w \sim N} \left[ R_k( \text{Sim}(C_{verified}, \theta) ) \right] $$
where $\text{Sim}(\cdot)$ is the simulation operator containing the system and parameters.

After each simulation closed loop, the system calculates the performance metrics $P^*_k$ and updates the historical memory $H_{k+1} = H_k \cup (\Phi_k, P^*_k)$. This drives the next round of meta-configuration refinement by the LLM, achieving automation and iterative refinement in parameter tuning.

To completely illustrate the dynamic optimization process of the PTA driven by simulation, we provide the complete pseudo-code for the double-layer iteration in Algorithm \ref{alg:tuning}.

\begin{algorithm}
\caption{Agent3 Tuning DoubleLoop}
\label{alg:tuning}
\algrenewcommand\algorithmicindent{2.0em} 
\begin{algorithmic}[1]
\Require 
    \Statex $C$: Controller
    \Statex $S_{spec}$: Target specifications
    \Statex $K$: Max agent refine round 
    \Statex $N$: Max simulations per round
\Ensure 
    \Statex $\theta^*$: Best parameters
    \Statex $J^*$: Best score
\State $\theta^* \gets \text{None};\ J^* \gets -\infty;\ F \gets \emptyset;\ H_{refine} \gets \emptyset$
\Statex \Comment{$F$ is the feedback from LLM refine histroy}
\For{$k = 1 \dots K$}
    \State $(R_k, \Omega_k) \gets \textsc{LlmDesignOrRefine}(C, S_{spec}, F)$
    \State $H_{optimizer} \gets \emptyset$
    \For{$t = 1 \dots N$}
        \State $\theta \gets \textsc{OptimizerSuggest}(\Omega_k, H_{optimizer}, R_k)$
        \State $(ok, A) \gets \textsc{CompilerSimulate}(\theta, C)$
        \If{\textbf{not} $ok$}
            \State $C \gets \textsc{DesignerDebug}(C, A)$
            \State \textbf{continue}
        \EndIf
        \State $M \gets \textsc{EvaluatePerformance}(A)$
        \State $J \gets \textsc{CalculateScore}(M)$
        \State $H_{optimizer} \gets \textsc{Update}(H_{optimizer}, \theta, M, J)$
        \State $H_{refine} \gets \textsc{Merge}(H_{refine}, \theta, M, J, k)$
        \If{$J > J^*$}
            \State $(\theta^*, J^*) \gets \textsc{BestUpdate}(\theta^*, J^*, \theta, J)$ 
        \EndIf
        \If{$\textsc{MeetTarget}(M, S_{spec})$}
            \State \Return $(\theta^*, J^*, H_{refine})$
        \EndIf
    \EndFor
    \State $F \gets \textsc{BuildFeedback}(H_{optimizer}, C)$
\EndFor
\State \Return $(\theta^*, J^*, H_{refine})$
\end{algorithmic}
\end{algorithm}

\begin{table*}[t] 
\centering
\caption{Comparison of passing rates (\%) among different control design methods across various dynamic systems. Our proposed agentic system consistently outperforms baseline methods, particularly in complex unstable and higher-order systems.}
\label{tab:method_comparison}
\setlength{\tabcolsep}{1.5pt} 
\begin{tabular}{l cccccccccccc}
\toprule
\multirow{2}{*}{\textbf{Methods}} & \multicolumn{2}{c}{\textbf{First-ord Stable}} & \multicolumn{2}{c}{\textbf{Second-ord Stable}} & \multicolumn{2}{c}{\textbf{First-ord Unstable}} & \multicolumn{2}{c}{\textbf{Second-ord Unstable}} & \multicolumn{2}{c}{\textbf{With Delay}} & \multicolumn{2}{c}{\textbf{Higher Order}} \\
\cmidrule(lr){2-3} \cmidrule(lr){4-5} \cmidrule(lr){6-7} \cmidrule(lr){8-9} \cmidrule(lr){10-11} \cmidrule(lr){12-13}
 & Rate & Iter. & Rate & Iter. & Rate & Iter. & Rate & Iter. & Rate & Iter. & Rate & Iter. \\
\midrule
AICS                    & \textbf{98\%} & \textbf{1.20} & \textbf{98\%} & \textbf{1.45} & \textbf{100\%} & \textbf{1.00} & \textbf{90\%} & \textbf{2.44} & \textbf{100\%} & \textbf{1.00} & \textbf{74\%} & 2.32 \\
ControlAgent            & \textbf{98\%} & 1.96 & \textbf{98\%} & 2.03 & 98\%  & 3.17 & 74\% & 6.83 & 96\%  & 1.06 & 54\% & \textbf{2.00} \\
Zero-shot               & 30\% & / & 0\% & / & 22\%  & / & 0\% & / & 62\%  & / & 8\% & / \\
Zero-shot with feedback & 70\% & 2.69 & 4\% & 6.50 & 50\%  & 2.92 & 2\% & 5.00 & \textbf{100\%}  & 1.78 & 20\% & 3.70 \\
Few-shot               & 30\% & / & 6\% & / & 10\%  & / & 0\% & / & 18\%  & / & 4\% & / \\
Few-shot with feedback & 96\% & 2.23 & 64\% & 5.19 & 76\%  & 4.45 & 26\% & 6.77 & \textbf{100\%}  & 2.14 & 40\% & 3.05 \\
PIDtune                 & 56\% & / & 82\% & / & 32\%  & / & 12\% & / & \textbf{100\%}  & / & 50\% & / \\
\bottomrule
\end{tabular}
\end{table*}

\begin{figure*}[htbp]
    \centering
    \includegraphics[width=\textwidth]{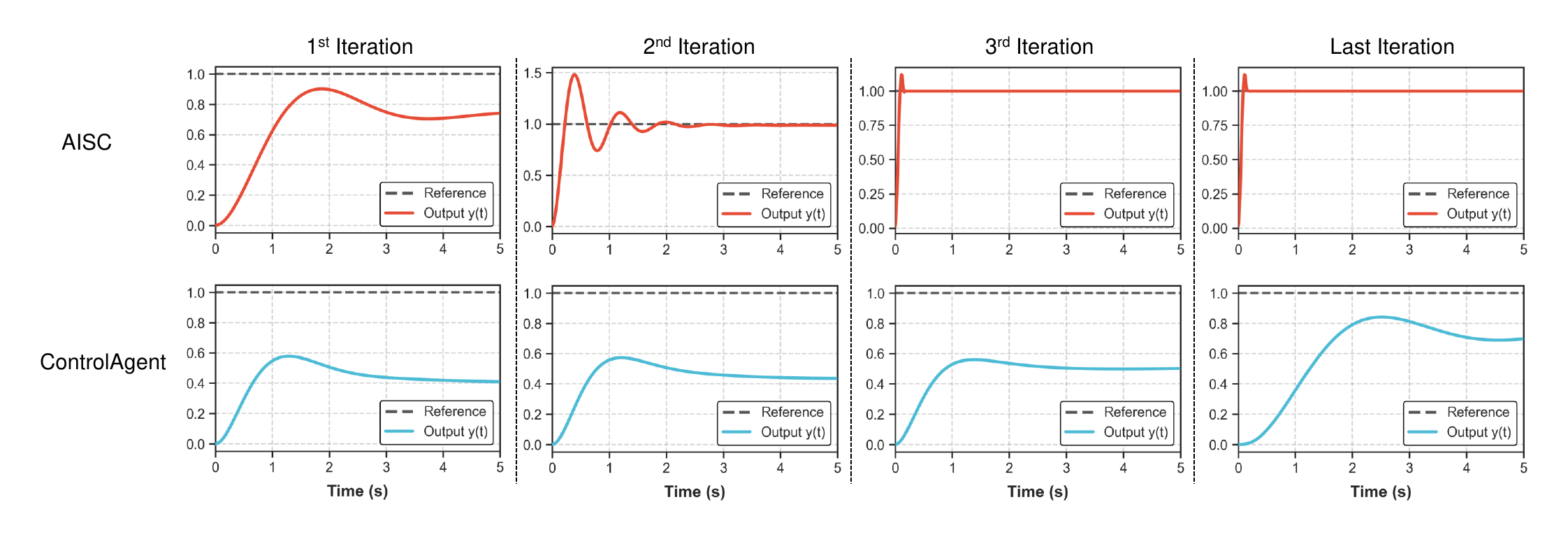} 
    \caption{Time-response evolution during the parameter tuning process. The proposed system (AICS, top row) rapidly explores the parameter space and converges to perfect reference tracking, whereas the baseline (ControlAgent, bottom row) gets trapped in a local optimum with severe steady-state error.}
    \label{fig:response_comparison}
\end{figure*}

\section{Experiments}
To comprehensively evaluate the effectiveness of the proposed agentic system (AICS), we designed a series of comprehensive experiments.

\subsection{Experimental Setup}
\textbf{Benchmark:} The test benchmark in this paper is built upon the existing ControlEval dataset, with an escalated difficulty level tailored to the characteristics of complex multi-agent collaborative tasks. The test systems include: first-order/second-order stable systems, first-order/second-order unstable systems, systems with pure time delay, and higher-order complex systems. Each category contains 50 independent test cases.

\textbf{Evaluation Metrics:} The experiments employ two core metrics for evaluation:
1) \textbf{Pass Rate:} Measures whether the generated controller successfully achieves the control objectives while satisfying all performance metrics and constraints.
2) \textbf{Number of Iterations:} Records the average number of refinement iterations required for the system to reach the optimal parameter configuration, serving as a measure of the search efficiency of the automated design.

\textbf{Model Configurations:} To ensure a fair comparison, the default LLM backbone utilized in our system ($\text{AICS}$) across all baseline experiments is Gemini-2.0-Flash. 

\begin{figure}[htbp]
    \centering
    \includegraphics[width=\linewidth]{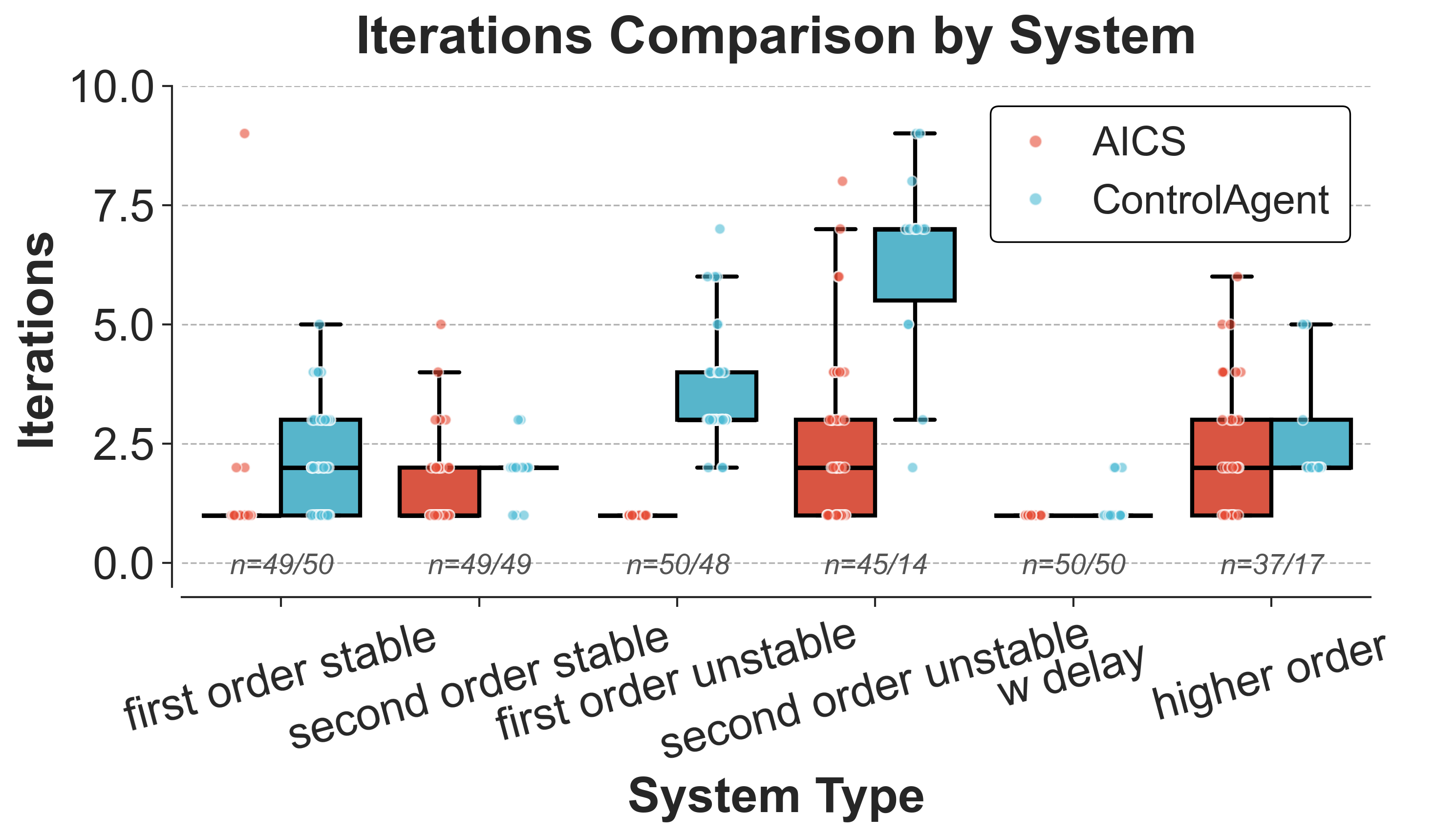} 
    \caption{Boxplot of iteration counts for successful trials. Ours demonstrates significantly lower medians, indicating faster and more stable convergence.}
    \label{fig:iterations_box}
\end{figure}

\subsection{Main Results and Baseline Comparison}
We compared the proposed system with two baseline methods: a traditional heuristic tuning tool (PIDtune) and a recently proposed LLM-based method (ControlAgent).


\begin{table*}[t] 
\centering
\caption{Performance comparison of different LLM backbones across various dynamic systems (RQ3). The table records the final pass rate (\%) within 10 iterations and the average number of iterations (Iter.) for successful samples.}
\label{tab:backbone_comparison}
\setlength{\tabcolsep}{1pt} 
\begin{tabular}{lc cccccccccccc}
\toprule
\multirow{2}{*}{\textbf{Model Backbone}} & \multirow{2}{*}{\textbf{Params}} & \multicolumn{2}{c}{\textbf{First-ord Stable}} & \multicolumn{2}{c}{\textbf{Second-ord Stable}} & \multicolumn{2}{c}{\textbf{First-ord Unstable}} & \multicolumn{2}{c}{\textbf{Second-ord Unstable}} & \multicolumn{2}{c}{\textbf{With Delay}} & \multicolumn{2}{c}{\textbf{Higher Order}} \\
\cmidrule(lr){3-4} \cmidrule(lr){5-6} \cmidrule(lr){7-8} \cmidrule(lr){9-10} \cmidrule(lr){11-12} \cmidrule(lr){13-14}
 & & Rate & Iter. & Rate & Iter. & Rate & Iter. & Rate & Iter. & Rate & Iter. & Rate & Iter. \\
\midrule
Qwen3-8B              & 8B       & 98\% & 1.39 & 98\% & 2.80 & 98\%  & 1.43 & 74\% & 2.24 & 96\%  & 1.12 & 54\% & 2.07 \\
GPT-4o-mini           & $\sim$8B & 90\% & 1.07 & 96\% & 1.92 & 96\%  & \textbf{1.00} & 82\% & 2.56 & 96\%  & 1.02 & 68\% & 3.44 \\
\cmidrule{1-14} 
Qwen3-32B             & 32B      & 96\% & 1.19 & \textbf{100\%} & 3.20 & 96\%  & 1.04 & 84\% & 2.31 & 92\%  & 1.04 & 66\% & 2.48 \\
GLM-4.7-Flash         & 30B      & 94\% & 1.17 & 92\% & 1.65 & 88\%  & 1.09 & 86\% & \textbf{1.72} & 92\%  & 1.07 & 70\% & 2.23 \\
\cmidrule{1-14} 
Qwen3-235B-A22B       & 235B     & \textbf{100\%} & 1.66 & 98\% & 2.06 & \textbf{100\%}  & 1.66 & \textbf{98\%} & 1.98 & 98\%  & 1.33 & \textbf{74\%} & 3.38 \\
Gemini-2.0-Flash      & -        & 98\% & 1.20 & 98\% & \textbf{1.45} & \textbf{100\%} & \textbf{1.00} & 90\% & 2.44 & \textbf{100\%} & \textbf{1.00} & \textbf{74\%} & 2.32 \\
Gemini-2.5-Flash      & -        & 98\% & \textbf{1.06} & 98\% & 2.06 & \textbf{100\%}  & 1.06 & 92\% & 2.28 & 98\%  & \textbf{1.00} & 66\% & \textbf{2.06} \\
\bottomrule
\end{tabular}
\end{table*}

The quantitative results across various dynamic systems are consolidated in Table~\ref{tab:method_comparison}. 
For relatively simple scenarios, such as first-order stable and time-delay systems, our system, \textit{ControlAgent}, and the feedback-enabled LLM baselines (i.e., Zero-shot and Few-shot with feedback) all yield near-perfect pass rates, whereas the traditional analytical method (\textit{PIDtune}) exhibits inconsistent performance. 

However, as the system complexity scales up, the profound limitations of open-loop LLM paradigms and traditional heuristics are rapidly exposed. 
Without structural prompt engineering or iterative execution guidance, the vanilla \textit{Zero-shot} and \textit{Few-shot} methods suffer from severe performance degeneration, failing completely (0\% pass rate) in demanding second-order unstable environments. 
Although introducing environment feedback noticeably boosts their tracking capabilities, their blind numerical exploration heavily compromises efficiency, leading to exorbitant iteration costs (e.g., a median of 6.50 iterations for \textit{Zero-shot with feedback} in second-order stable systems). 

When confronted with highly nonlinear second-order unstable systems, \textit{PIDtune} severely diverges with a pass rate of only 12\%, while \textit{ControlAgent}—which lacks a coupled mathematical optimization backbone—caps at 74\%. 
In stark contrast, our proposed agentic system maintains a robust pass rate of 90\% with a significantly minimized optimization budget (2.44 iterations). 
A similar superior trend is observed in higher-order complex systems, where our system achieves a 74\% pass rate, vastly outperforming \textit{ControlAgent} (54\%), \textit{Few-shot with feedback} (40\%), and \textit{PIDtune} (50\%). 
This firmly substantiates the necessity and superior capability of our system in decoupling symbolic reasoning from precise numerical computations for intricate control synthesis.

The boxplot in Fig.~\ref{fig:iterations_box} further reveals the advantage of the multi-agent collaborative mechanism in search efficiency. 
This figure intuitively illustrates the statistical distribution of iteration counts when models successfully achieve the control targets. 
It can be clearly observed that when tackling challenging second-order unstable systems, the median iteration count for \textit{ControlAgent} reaches around 6, with its upper quartile at 7 and long-tail outliers extending up to 9 iterations. 
This indicates significant blindness and randomness in the parameters generated by LLM when confronted with complex systems. 
Conversely, our system (AICS) demonstrates a clear advantage in iteration counts across all system types, proving its high optimization efficiency.

The dynamic response comparison in Fig.~\ref{fig:response_comparison} illustrates the performance discrepancy between our method and \textit{ControlAgent} on the same system. 
Compared to \textit{ControlAgent}, the proposed method can more accurately capture performance flaws fed back from historical information during iterations, and swiftly execute corrections.

\begin{figure}[htbp]
    \centering
    \includegraphics[width=\linewidth]{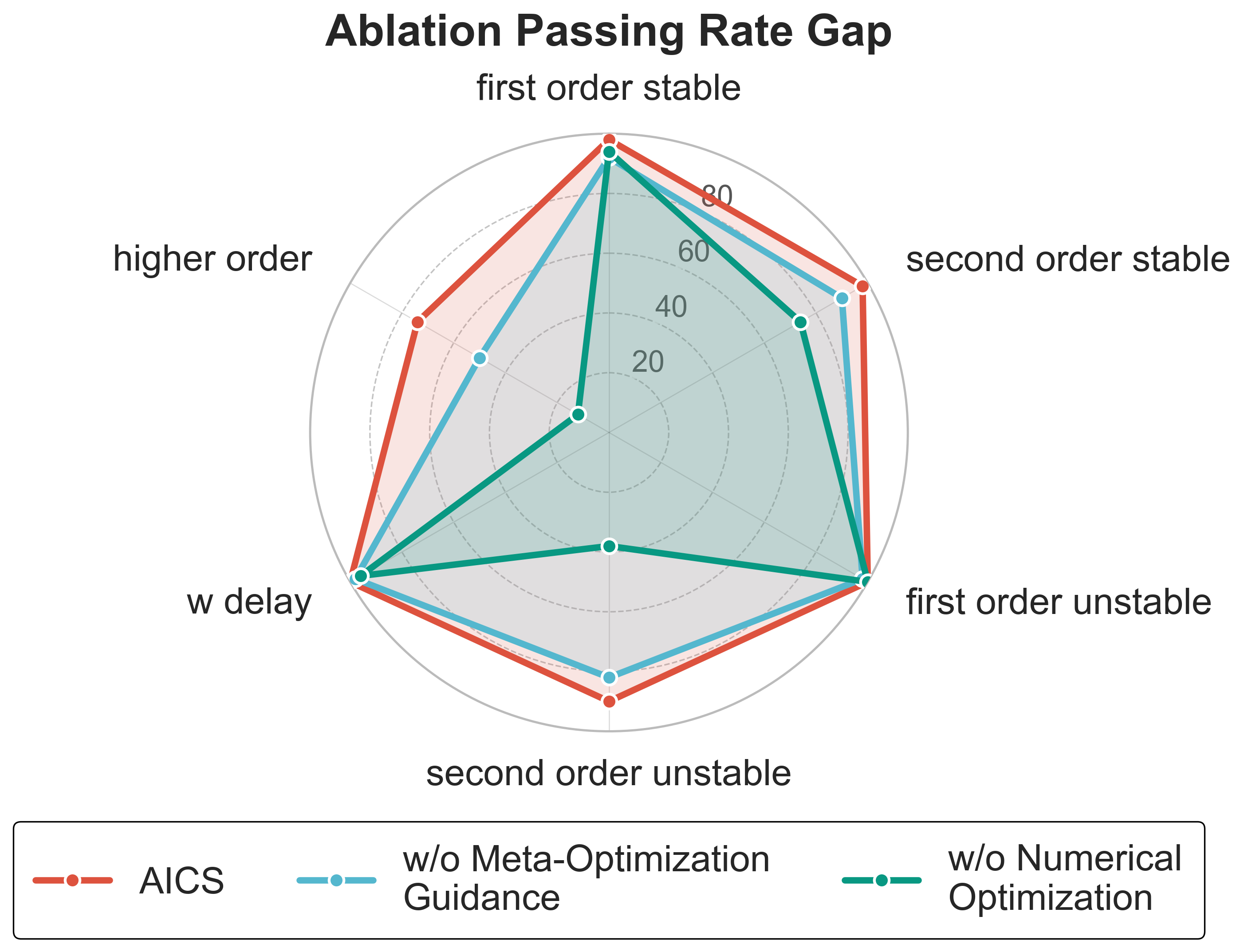}
    \caption{Ablation passing rate gap across different system types. The area collapse of the green line illustrates the critical role of numerical optimization in complex systems.}
    \label{fig:ablation_radar}
\end{figure}

\begin{figure*}
    \centering
    \includegraphics[width=\textwidth]{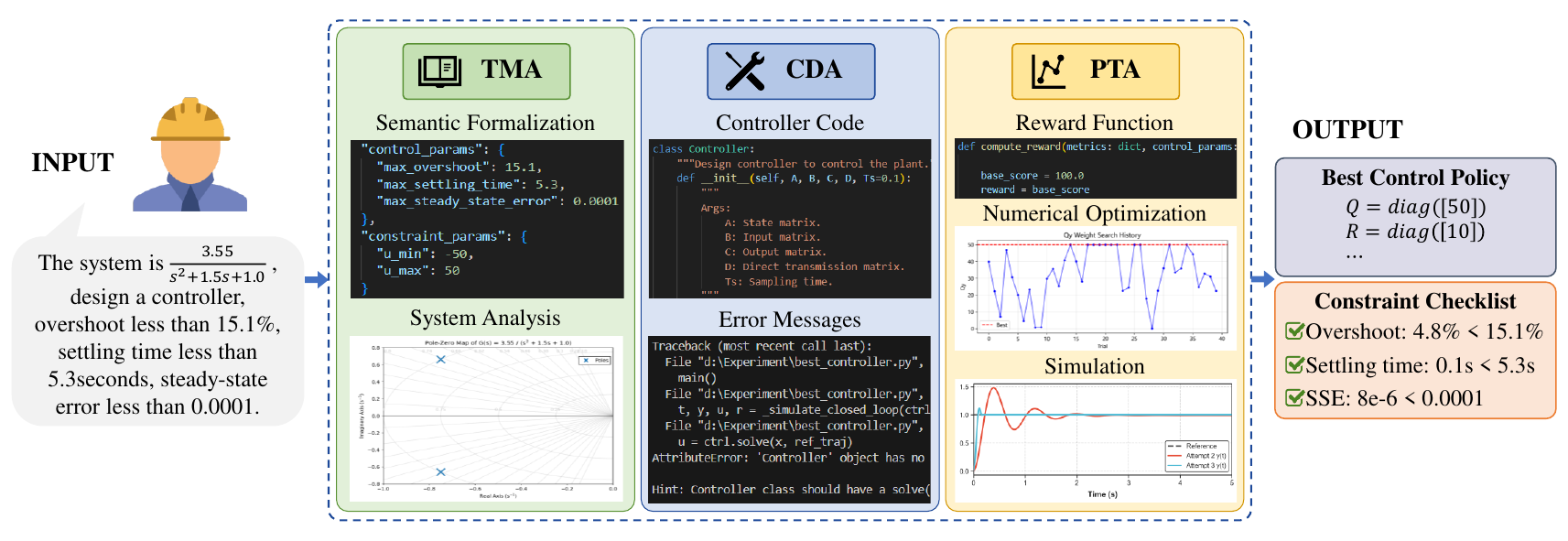} 
    \caption{A comprehensive case study of the proposed system. It illustrates the transparent and interpretable workflow across Semantic Formalization (Stage 1), Code Generation with error self-correction (Stage 2), and Simulation-driven Parameter Tuning (Stage 3).}
    \label{fig:case_study}
\end{figure*}

\begin{figure}[htbp]
    \centering
    \includegraphics[width=\linewidth]{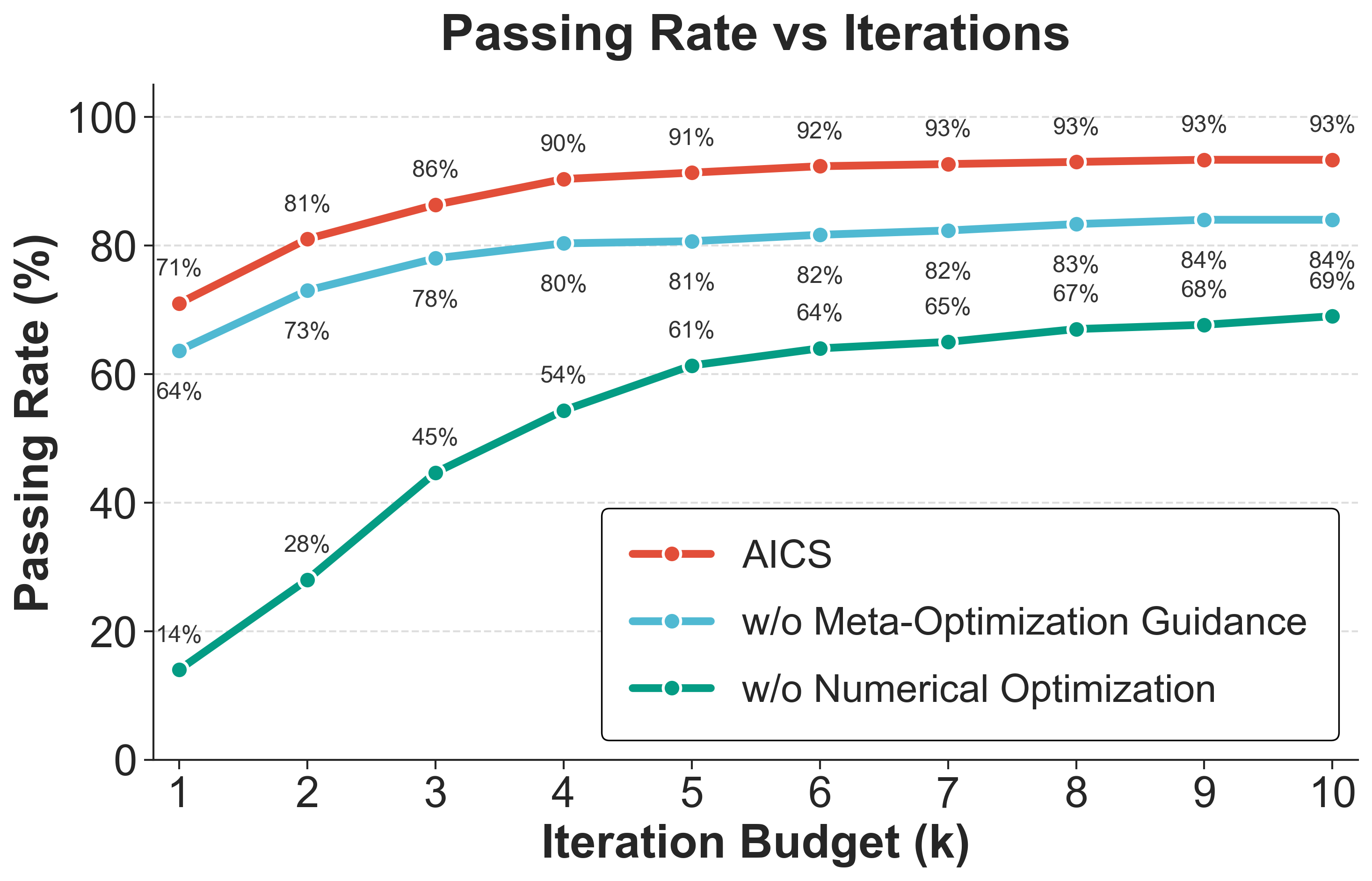}
    \caption{Passing rate versus iteration budget. The proposed method (Ours) achieves faster convergence and a higher performance upper bound compared to the ablated variants.}
    \label{fig:ablation_curve}
\end{figure}

\subsection{Ablation Study on Multi-Agent Mechanisms}
To verify the necessity of key components within the system, we designed ablation studies by removing the "Numerical Optimization module (w/o Numerical Optimization)" and the "LLM Historical Feedback Guidance module (w/o Meta-Optimization Guidance)" respectively.

\textbf{1. The Necessity of Numerical Optimization:} 
As illustrated by the radar chart in Fig.~\ref{fig:ablation_radar}, when the system degrades to a mode relying solely on the LLM for direct parameter generation (w/o Numerical Optimization), its performance envelope exhibits a severe "area collapse." Particularly when facing higher-order and second-order unstable systems, the pass rate shrinks drastically. Combined with the convergence curves in Fig.~\ref{fig:ablation_curve}, it is evident that the initial success rate of this ablated version is extremely low. Furthermore, as the iteration budget increases, its convergence is slow, and its final performance upper bound remains lower than the variants equipped with numerical optimizers. This result experimentally validates our core hypothesis: LLMs are unreliable when directly handling rigorous engineering numerical computations. The "decoupling of symbolic reasoning and numerical computation" strategy adopted by this system is a critical safeguard for complex control design.

\textbf{2. The Impact of Meta-Optimization Guidance:} 
Fig.~\ref{fig:ablation_curve} reveals the accelerating effect of the LLM historical feedback guidance module on search efficiency. When the heuristic guidance based on Trial History is removed (blue curve), although the system eventually reaches an acceptable overall pass rate, its early convergence speed lags noticeably behind the full system (red curve), and its final asymptotic upper bound (approximately $84\%$) consistently fails to approach the extremum of the full system (over $90\%$). Moreover, correlating this with Fig.~\ref{fig:ablation_radar} reveals that the absence of this module leads to a noticeable performance degradation specifically in challenging scenarios like second-order unstable and higher-order systems. This indicates that without meta-optimization guidance, the parameter configurations provided by the LLM easily fall into blind searches or local optima, struggling with demanding control tasks. The LLM's meta-reasoning capability, grounded in historical data, can effectively and dynamically narrow the parameter search space $\Omega_k$, assisting the system in quickly locking onto the global optimal solution with a smaller Iteration Budget.

\subsection{Impact of Model Scale and Reasoning Capabilities}
Finally, under the same multi-agent architecture, we substituted different underlying LLM backbones, encompassing open-source and cutting-edge closed-source models with parameter scales ranging from 8B to 235B, to investigate the impact of model capabilities on control system design.

Observing the data in Table~\ref{tab:backbone_comparison}, this system demonstrates excellent model generalizability. For delay systems and first-order systems, smaller parameter models are already competent for the task, providing pass rates comparable to ultra-large models. However, there is a distinct correlation between the pass rate on complex tasks and model reasoning capabilities. When handling second-order unstable systems and higher-order systems, large models equipped with stronger logic chains and code synthesis abilities exhibit superior performance. For instance, Qwen3-235B achieves a $74\%$ pass rate on higher-order tasks, consistent with our default configuration, while the 8B model only achieves $54\%$. This indicates that state-space modeling and reward function reconstruction for complex control systems are deeply tied to the underlying backbone's semantic understanding capability. As the reasoning capabilities of future LLMs continue to evolve, the automated design ceiling of this system will be continuously expanded.

\subsection{Case Study: End-to-End Autonomous Design}
To more intuitively illustrate the reasoning and and self-correction capabilities of our system, we analyze a representative case study of a second-order plant
$$G(s) = \frac{3.55}{s^2 + 1.5s + 1.0}$$
 As shown in Fig.~\ref{fig:case_study}, the end-to-end execution progresses through three collaborative stages:
\begin{itemize}
    \item \textbf{Semantic Formalization:} The TMA eliminates linguistic ambiguity by mapping the textual constraints into deterministic parameters such as \texttt{"max\_overshoot": 15.1}, \texttt{"max\_settling\_time": 5.3}, then performs open-loop system analysis to establish a rigorous mathematical feasible region.
    \item \textbf{Code Generation:} The CDA generates the controller code. When the initial code triggers an execution exception, the CDA automatically captures the traceback stack and error messages, utilizing its control theory verification engine to refactor the logic in real-time without human intervention.
    \item \textbf{Parameter Tuning:} To bypass numerical hallucinations, the PTA decouples symbolic reasoning from calculation. The LLM acts as a planner to configure the reward function and search boundaries, while the underlying optimizer drives closed-loop simulations to lock onto the optimal parameters $Q = \text{diag}([50])$ and $R = \text{diag}([10])$.
\end{itemize}
\textbf{Output Verification:} The final checklist validated by the simulation environment demonstrates that the generated controller satisfies all industrial criteria, achieving an overshoot of $4.8\% < 15.1\%$, a settling time of $0.1\,\text{s} < 5.3\,\text{s}$, and a steady-state error of $8 \times 10^{-6} < 0.0001$. This case study validates the system's high reliability, transparency, and closed-loop robustness for autonomous control engineering.

\bibliographystyle{ieeetr}
\bibliography{reference}

\end{document}